%% file: main.tex
\pdfoutput=1

\documentclass[11pt]{article}

\usepackage[]{acl}

\usepackage{times}
\usepackage{latexsym}

\usepackage[T1]{fontenc}

\usepackage[utf8]{inputenc}

\usepackage{microtype}

\title{User-Centric Gender Rewriting}

\author{Bashar Alhafni, Nizar Habash, Houda Bouamor\textsuperscript{\textdagger}\\
  Computational Approaches to Modeling Language Lab\\
  New York University Abu Dhabi\\
  \textsuperscript{\textdagger}Carnegie Mellon University in Qatar\\
  \texttt{\{alhafni,nizar.habash\}@nyu.edu},
  \texttt{hbouamor@qatar.cmu.edu}
  }
  
\usepackage{graphicx}
\usepackage{url}
\usepackage{latexsym}
\usepackage{makecell}
\usepackage{epstopdf}
\usepackage{tipa}
\usepackage{hyperref}
\usepackage{xstring}
\usepackage{amsfonts}
\usepackage{amsmath}
\usepackage{fixltx2e}
\usepackage[T1]{fontenc}
\usepackage{booktabs}
\usepackage{arabtex}
\usepackage{hhline}
\usepackage{pdfpages}
\usepackage{utf8}
\input{macros.tex}
\usepackage{multirow}
\usepackage{subfigure}
\usepackage{amssymb}
\usepackage[super]{nth}

\usepackage{caption}
\setcode{utf8}
\vocalize
\usepackage{xcolor}

\definecolor{raspberry}{HTML}{eb0170}
\definecolor{purple(x11)}{HTML}{9c00fe}

\newcommand{\APGC}{{\sc APGC}}

\newcommand{\train}{{\sc Train}}
\newcommand{\dev}{{\sc Dev}}
\newcommand{\test}{{\sc Test}}

\begin{document}
\maketitle
\begin{abstract}
In this paper, we define the task of gender rewriting in contexts involving two users (I~and/or~You) -- first and second grammatical persons with independent grammatical gender preferences. We focus on Arabic, a gender-marking morphologically rich language. We develop a multi-step system that combines the positive aspects of both rule-based and neural rewriting models. Our results successfully demonstrate the viability of this approach on a recently created corpus for Arabic gender rewriting, achieving 88.42 M\textsuperscript{2} F\textsubscript{0.5} on a blind test set. Our proposed system improves over previous work on the first-person-only version of this task, by 3.05 absolute increase in M\textsuperscript{2} F\textsubscript{0.5}. We demonstrate a use case of our gender rewriting system by using it to post-edit the output of a commercial MT system to provide personalized outputs based on the users' grammatical gender preferences. We make our code, data, and models publicly available.\footnote{\url{https://github.com/CAMeL-Lab/gender-rewriting/}}

\end{list}
\end{abstract}

\section{Introduction}

Gender bias is a fundamental problem in natural language processing (NLP) and it has been receiving an increasing attention across a variety of core tasks such as machine translation (MT), co-reference resolution, and dialogue systems. Research has shown that NLP systems have the ability to embed and amplify gender bias \cite{sun-etal-2019-mitigating}, which not only degrades users’ experiences but also creates representational harm \cite{blodgett-etal-2020-language}. The embedded bias within NLP systems is usually attributed to training models on biased data that reflects the social inequalities of the world we live in. However, even the most balanced of models can still exhibit and amplify bias if they are designed to produce a single text output without taking their users' gender preferences into consideration. Therefore, to provide the correct user-aware output, NLP systems should be designed to
produce outputs that are as gender-specific as the users information they have access to. Users information could be either embedded as part of the input or provided externally by the users themselves. 
In cases where this information is unavailable to the system, generating all gender-specific forms or a gender-neutral form is more appropriate.

Producing user-aware outputs becomes more challenging for systems targeting multi-user contexts (first and second persons, with independent grammatical gender preferences), particularly when dealing with gender-marking morphologically rich languages.
In this paper, we define the task of gender rewriting in contexts involving two users (I and/or You) -- first and second grammatical persons with independent grammatical gender preferences and we focus on Arabic, a gender-marking morphologically rich language.  The main contributions of our work are as follows: 

\begin{enumerate}
    \item We introduce a multi-step gender rewriting system that combines the positive aspects of rule-based and neural models.

  \item We demonstrate our approach's effectiveness by establishing a strong benchmark on a publicly available multi-user Arabic gender rewriting corpus.

  \item We show that our best system yields state-of-the-art results on the first-person-only version of this task, beating previous work.

  \item We demonstrate a use case of our system by post-editing the output of an MT system to match users' grammatical gender preferences.

\end{enumerate}





This paper is organized as follows. We first discuss related work (\S\ref{sec:background}) as well as relevant Arabic linguistic facts (\S\ref{sec:arabic-facts}). 
We then define the gender rewriting task in  \S\ref{sec:task} and describe the data we use and the gender rewriting model we build in \S\ref{sec:data_background} and \S\ref{sec:model_desc}. Lastly, we present our experimental setup (\S\ref{sec:experiments}) and results (\S\ref{sec:results}) and conclude in \S\ref{sec:conclusion}.


\section{Background and Related Work}
\label{sec:background}
Substantial research has targeted 
the problem of gender bias in various NLP
tasks such as MT
\cite{rabinovich-etal-2017-personalized,vanmassenhove-etal-2018-getting,stafanovics2020mitigating,savoldi2021gender}, 
dialogue systems \cite{cercas-curry-etal-2020-conversational,dinan-etal-2020-queens,liu-etal-2020-gender,liu-etal-2020-mitigating,sheng2021revealing}, 
language modeling \cite{lu2018gender,bordia-bowman-2019-identifying,sheng-etal-2019-woman,vig2020,nadeem-etal-2021-stereoset}, 
co-reference resolution \cite{rudinger-etal-2018-gender,zhao-etal-2018-gender}, and named entity recognition \cite{mehrabi2019man}. 
While the majority of research has focused on tackling gender bias in English by debiasing word embeddings \cite{bolukbasi2016man,zhao-etal-2018-learning,gonen2019lipstick,manzini-etal-2019-black,zhao2020gender} or by training systems on gender-balanced corpora built using counterfactual data augmentation techniques \cite{lu2018gender,hall-maudslay-etal-2019-name,zmigrod-etal-2019-counterfactual}, our work falls under text rewriting through the controlled generation of gender alternatives for morphologically rich languages. 

Within text rewriting, \newcite{vanmassenhove-etal-2021-neutral} and \newcite{sun2021they} recently presented rule-based and neural rewriting models to generate gender-neutral sentences in English. For morphologically rich languages and specifically Arabic, \newcite{habash-etal-2019-automatic} introduced the Arabic Parallel Gender Corpus v1.0 ({\APGC}~v1.0) of first-person-singular constructions and designed a two-step gender identification and reinflection system to generate masculine and feminine grammatical gender alternatives. 
\newcite{alhafni-etal-2020-gender} used {\APGC}~v1.0
to create a joint gender identification and reinflection model. They treated the
problem as a user-aware grammatical error correction task and showed improvements over \newcite{habash-etal-2019-automatic}'s results. Both efforts modeled gender reinflection using character-level Seq2Seq models. More recently, \newcite{Alhafni:2022:corpus} extended {\APGC}~v1.0 to~{\APGC}~v2.0 by including contexts involving first and second grammatical persons covering singular, dual, and plural constructions; and adding six times more sentences. 

In our work, we use {\APGC}~v2.0 to build a multi-step gender rewriting system to generate gender alternatives in multi-user contexts. We also  show improvements over both \newcite{habash-etal-2019-automatic}'s and \newcite{alhafni-etal-2020-gender}'s results on {\APGC}~v1.0.

\begin{table*}[t]
\centering
\includegraphics[width=0.95\linewidth]{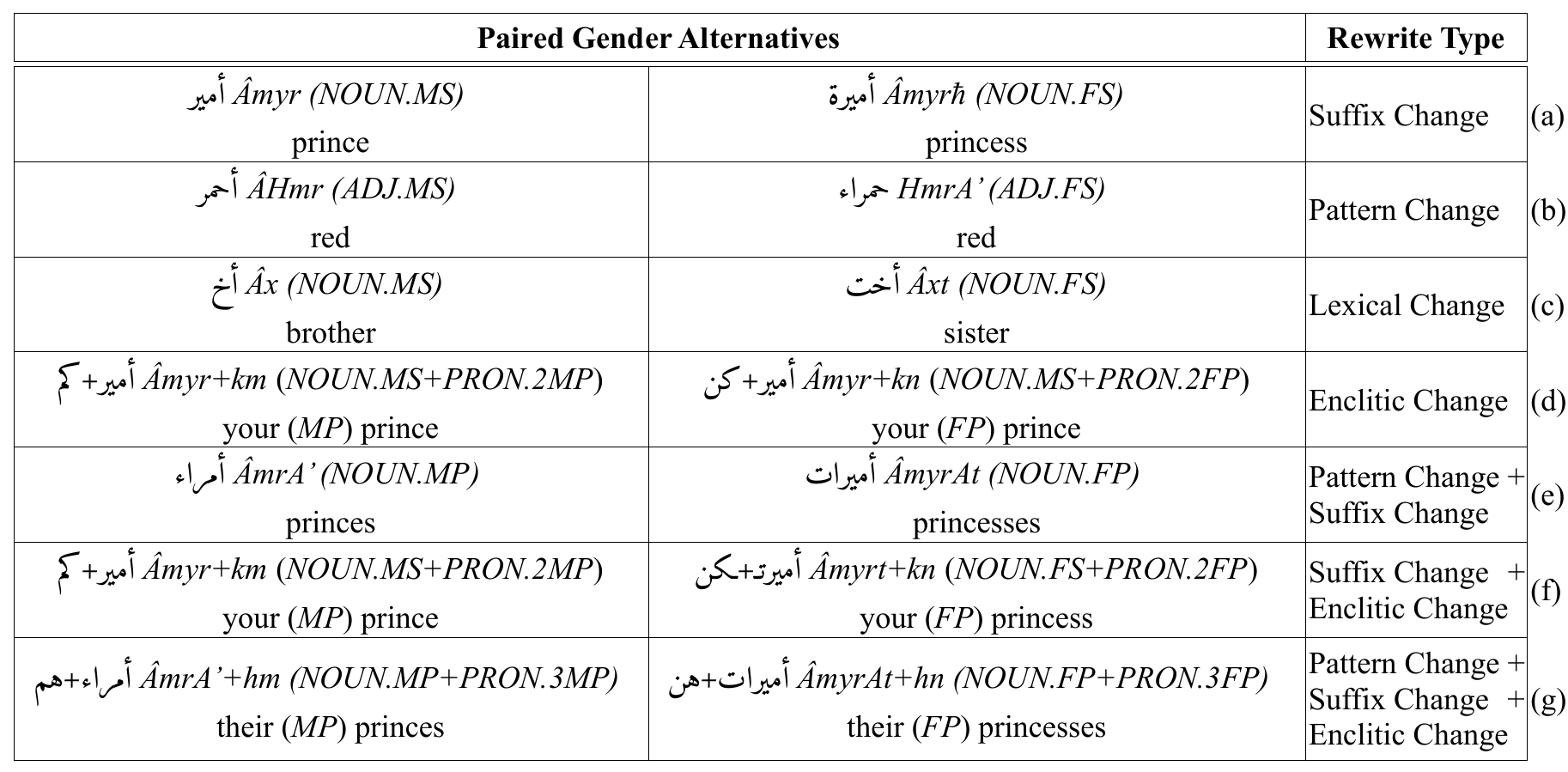}
\caption{Examples of the changes needed to generate gender alternative forms of gender-specific words in Arabic.}
\label{tab:gen_examples}
\end{table*}

\section{Arabic Linguistic Facts}
\label{sec:arabic-facts}

We highlight two of the many challenges that face Modern Standard Arabic (MSA) NLP systems dealing with gender expressions.

\paragraph{Morphological Richness and Complexity} Arabic has a rich and complex morphological system that inflects for many morphological features (gender, number, person, case, state, aspect, mood, voice), in addition to several attachable clitics (prepositions, particles, pronouns) \cite{Habash:2010:introduction}. 
Arabic nouns, adjectives,
and verbs inflect for gender: masculine (\textit{M}) and feminine (\textit{F}), and for number: singular (\textit{S}), dual (\textit{D}) and plural (\textit{P}). 

Changing the grammatical gender of Arabic words involves either changing the form of the \textit{base word}, changing the \textit{pronominal enclitics} that are attached to the \textit{base word}, or a combination of both. A \textit{base word} in Arabic refers to the stem along with its attachable affixes (prefixes, suffixes, circumfixes). Changing the \textit{base word} gender requires either a suffix change, a pattern change, or a lexical change as shown in Table~\ref{tab:gen_examples}(a-c). Arabic also has clitics that attach to the stem after affixes. A clitic is a morpheme that has the syntactic characteristics of a word but shows evidence of being phonologically bound to another word. In this respect, a clitic is distinctly different from an affix, which is phonologically and syntactically part of the word. Proclitics are clitics that precede the word (like a prefix), whereas enclitics are clitics that follow the word (like a suffix). \textit{Pronominal enclitics} are pronouns that cliticize to previous words (Table~\ref{tab:gen_examples}(d)). It is worth noting that multiple affixes and clitics can appear in a single word in Arabic and changing the grammatical gender of such words requires changing the genders of both the base word and its clitics (Table~\ref{tab:gen_examples}(f-g)).

\paragraph{Orthographic Ambiguity}
Arabic uses diacritics to specify short vowels. However, these optional diacritics are usually omitted in  Arabic orthography, leaving readers to infer the meaning of certain words based on the context \cite{Habash:2010:introduction}. This increases the degree of word ambiguity as gender-specific words could only differ in terms of diacritics. For instance, the verb \<لعبت>~\textit{l\AYN bt}\footnote{Arabic transliteration is in the HSB scheme \cite{Habash:2007:arabic-transliteration}.} can be diacritized as
\<لَعِبْتَ>
~\textit{la{\AYN}ibt{a}} `you [masc.] played' or as
\<لَعِبْتِ>
~\textit{la{\AYN}ibt{i}} `you [fem.] played'.


\section{The Gender Rewriting Task}
\label{sec:task}
We define the task of gender rewriting as generating alternatives of a given Arabic sentence to match different {\it target user gender contexts} (e.g., female speaker with a male listener, a male speaker with a male listener, etc.). This requires changing the grammatical gender (masculine or feminine) of certain words referring to the users (speaker/\nth{1} person and listener/\nth{2} person). Previous work done by \newcite{habash-etal-2019-automatic} and \newcite{alhafni-etal-2020-gender} refer to this task as gender reinflection, but we believe that gender rewriting is a more appropriate term given that it goes beyond reinflection.\footnote{Morphological reinflection usually refers to reinflecting either a lemma or an already inflected form to produce a desired form of a particular word~\cite{Cotterell:2016:sigmorphon,Cotterell:2017:conll-sigmorphon}.}


\paragraph{Notation} We will use four elementary symbols to facilitate the discussion of this task: 1M, 1F, 2M and 2F. The digit part of the symbol refers to the grammatical persons (\nth{1} or \nth{2}) and the letter part refers to the grammatical genders (masculine or feminine). Additionally, we will use B to refer to invariant/ambiguous gender.

We define the \textit{\textbf{sentence-level}} gender using the following four labels: 1M/2F, 1F/2M, 1M/2F, and 1F/2F. These four labels indicate the grammatical persons and genders of the user contexts we are modeling.

We define the \textit{\textbf{word-level}} gender based on the genders of the word's base form and its attachable pronominal enclitics (\S\ref{sec:arabic-facts}) using the notation: \textit{base form gender + enclitic gender}. This results in 25 word-level gender labels (e.g., B+1F, 1F+2M). We use B to refer to gender invariant/ambiguous words. Examples of the word-level gender labels are shown in Table~\ref{tab:data_examples}.

\paragraph{Task Definition} 
Given an Arabic sentence and a sentence-level target gender, the goal is to rewrite the input sentence to match the target users' gender preferences.

Some of the models we explore only use sentence-level gender labels; while other models use word-level gender labels to identify which input words need to be rewritten to match the target users' gender preferences.

\begin{table*}[t]
\centering 
\includegraphics[width=0.95\linewidth]{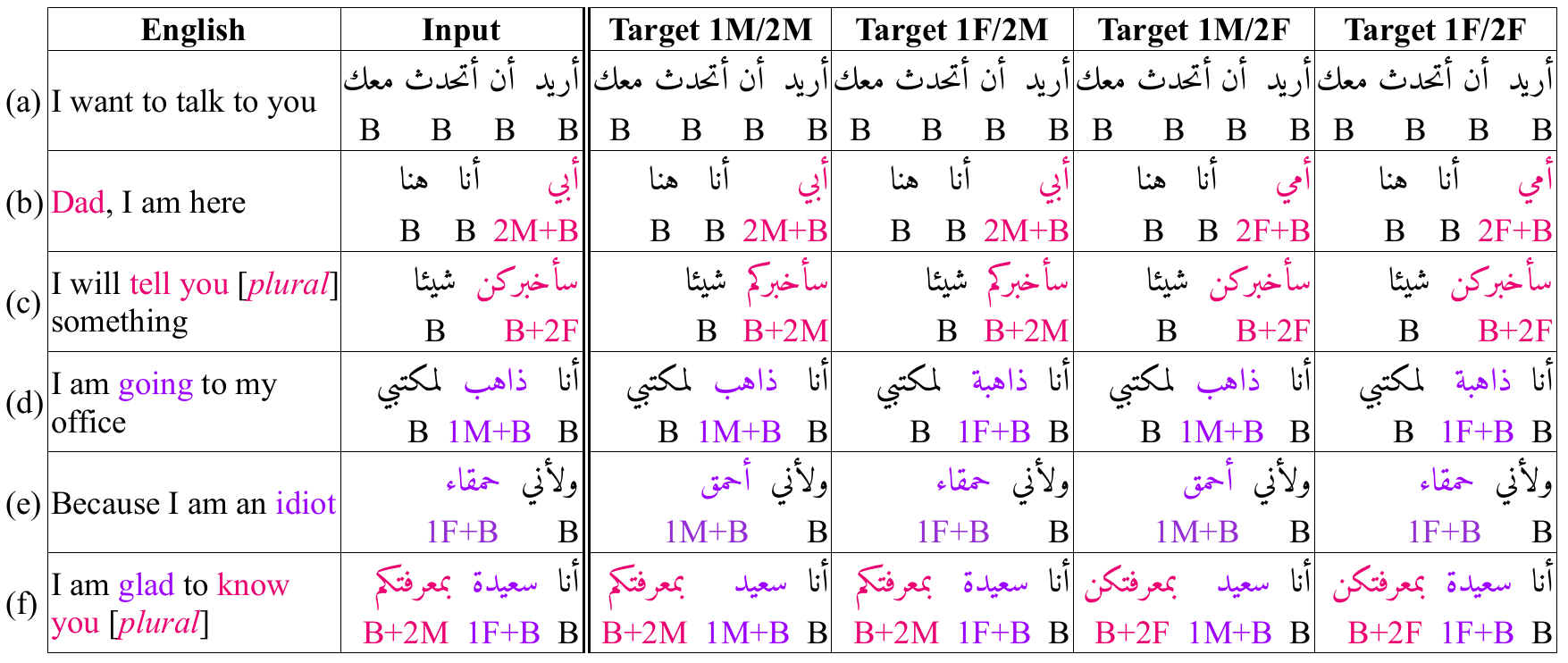}
\caption{Examples from the Arabic Parallel Gender Corpus v2.0 including the extended word-level annotations for each sentence and its rewrite to the opposite grammatical gender forms where appropriate. First person gendered words are in \textcolor{purple(x11)}{purple} and second person gendered words are in \textcolor{raspberry}{red}. M is Masculine; F is Feminine; and B is invariant.}
\label{tab:data_examples}
\end{table*}


\section{Data}
\label{sec:data_background}
For our experiments, we use the publicly available Arabic Parallel Gender Corpus (\APGC) -- a parallel corpus of Arabic sentences with gender annotations and gender rewritten alternatives of sentences selected from OpenSubtitles 2018~\cite{Lison:2016:opensubtitles2016}.
%
The corpus comes in two versions: {\APGC}~v1.0 and {\APGC}~v2.0. {\APGC}~v1.0 was introduced by \newcite{habash-etal-2019-automatic} and it contains 12,238 first-person-singular Arabic parallel gender-annotated sentences. \newcite{Alhafni:2022:corpus} expanded {\APGC}~v1.0 by including contexts involving first and second grammatical persons covering singular, dual, and plural constructions to create v2.0, which contains 80,326 gender-annotated parallel sentences (596,799 words). 
Both versions of {\APGC} include the original English parallels of the Arabic sentences.


In all of our experiments, we use an extended version of {\APGC}~v2.0 to train and test our systems. We also report results on the test set of {\APGC}~v1.0 to compare with previous work.

\paragraph{Annotations} Each sentence in {\APGC}~v2.0 has word-level gender labels where each word is labeled as 
B, 1F, 2F, 1M, or 2M.
All sentences containing gender-specific words referring to human participants have parallels representing their opposite gender forms. For the sentences without any gender-specific words, their parallels are trivial copies. Out of the 80,326 sentences in {\APGC}~v2.0, 46\% (36,980) do not contain any gendered words, whereas sentences with gendered references constitute 54\% (43,346). In terms of the word-level statistics, 9.7\% (58,066) are gender-specific, whereas 90.3\% (538,733) are marked as B.

Moreover, {\APGC}~v2.0 is organized into five parallel corpora that are fully aligned  (1-to-1) at the word level: Input, Target 1M/2M, Target 1F/2M, Target 1M/2F, and Target 1F/2F.
All five corpora are balanced in terms of gender, i.e., the number of 1F and 1M words is the same; and the number of 2F and 2M words is the same.
The Input corpus contains sentences with all possible word types (B, 1F, 2F, 1M, 2M). The Target 1M/2M corpus contains sentences that consist of B, 1M, 2M words; the Target 1F/2M corpus contains sentences that consist of B, 1F, 2M words; the Target 1M/2F corpus contains sentences that consist of B, 1M, 2F words; and the Target 1F/2F corpus contains sentences that consist of B, 1F, 2F words.

\paragraph{Splits} We use \newcite{Alhafni:2022:corpus}'s splits: 57,603 sentences (427,523 words) for training ({\train}), 6,647 sentences (49,257 words) for development ({\dev}), and 16,076 sentences (120,019 words) for testing ({\test}). 

\paragraph{Preprocessing the Word-Level Annotations}

Since gender information could be expressed at different parts of Arabic words (\S\ref{sec:arabic-facts}), we automatically extend the {\APGC}~v2.0 word-level annotations to mark the genders of both the base words and their pronominal enclitics.

Our preprocessing pipeline considers the labeled gendered words across the five parallel forms of each sentence in {\APGC}~v2.0. If the word ends with a gender marking pronominal enclitic, we label the gender of the enclitic based on predefined rules as 1F, 1M, 2F, or 2M. If the gendered word does not end with a gender-marking enclitic, then we label the enclitic as B. Once the enclitic is labeled, we compare the base form of the word across its parallel forms. If the base form is the same, we label it as B. Otherwise, we assign the base form the same label that is provided as part of {\APGC}~v2.0. All gender ambiguous words will be labeled as B. 
Table~\ref{tab:data_examples} presents some examples from {\APGC}~v2.0 with the extended word-level gender annotations. 

The extended word-level statistics are presented in Appendix~\ref{sec:extended_word_labels}. We make the extended word-level gender annotations publicly available as a new release of {\APGC} ({\APGC}~v2.1).\footnote{\url{http://resources.camel-lab.com/}}

\section{The Multi-step Model Approach}
\label{sec:model_desc}


Most of the recent work on gender rewriting rely on using Seq2Seq models \cite{habash-etal-2019-automatic,alhafni-etal-2020-gender,sun2021they,jain-etal-2021-generating,vanmassenhove-etal-2021-neutral}. 
However, the lack of large gender-annotated parallel datasets presents a challenge when training Seq2Seq models, and especially when dealing with morphologically rich languages.
This issue is highlighted by \newcite{alhafni-etal-2020-gender}, who report that most of the errors (68\%) produced by their character-level Seq2Seq model are due to not making any changes to gender-specific words in the input sentences. 
%
Given the complexity of the gender rewriting task in Arabic and the relatively small training data size,
we model the gender rewriting task using a multi-step system the combines the positive aspects of rule-based and neural models. Our system consists of three  components: \textit{Gender identification}, \textit{Out-of-context word gender rewriting}, and \textit{In-context ranking and selection}. 





\subsection{Gender Identification (GID)}
We first identify the word-level gender label (base word + pronominal enclitic) for each word in the input sentence. We build a word-level classifier by leveraging a Transformer-based pretrained language model.
There are many Arabic monolingual BERT models available such as AraBERT \cite{antoun-etal-2020-arabert}, ARBERT \cite{abdul-mageed-etal-2021-arbert}, and QARIB \cite{abdelali2021pretraining}. However, we chose to use CAMeLBERT MSA \cite{inoue-etal-2021-interplay} as it was pretrained on the largest MSA dataset to date.
Following the work of \newcite{devlin-etal-2019-bert}, we fine-tune CAMeLBERT MSA using Hugging Face's transformers \cite{wolf-etal-2020-transformers} by adding a fully-connected linear layer with a softmax on top of its architecture. During fine-tuning, we use the representation of the first sub-token as an input to the linear layer. 




\begin{figure}[t!]
\centering
\includegraphics[width=0.374\textwidth]{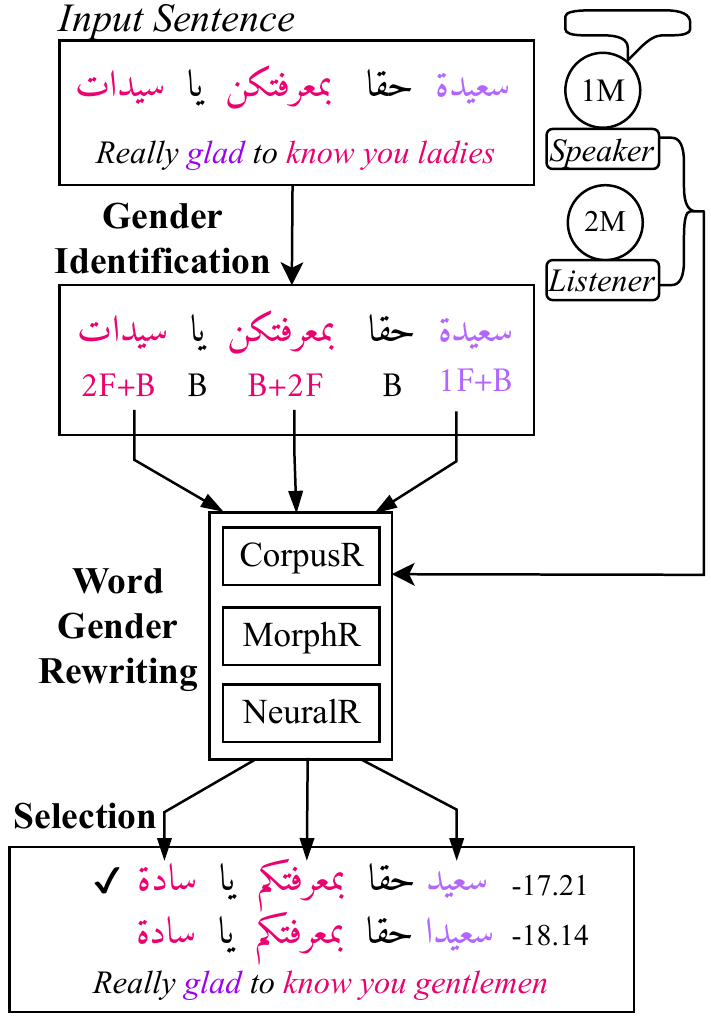}
\caption{The multi-step gender rewriting system. First person gendered words are in \textcolor{purple(x11)}{purple} and second person gendered words are in \textcolor{raspberry}{red}. The sentence-level target gender is 1M/2M. The input words \textit{glad} (1F+B), \textit{know you} (B+2F), and \textit{ladies} (2F+B) are rewritten to their masculine forms.}
\label{fig:model_diag}
\end{figure}

\subsection{Out-of-context Word Gender Rewriting}
Given the desired sentence-level target gender as an input and the identified gender label for each word in the input sentence, we decide if a word-level gender rewrite is needed based on the compatibility between the provided sentence-level target gender and the predicted word-level gender labels. 
%
We implement three word-level gender alternative generation models: \textit{Corpus-based Rewriter}, \textit{Morphological Rewriter}, and \textit{Neural Rewriter}. 

\paragraph{Corpus-based Rewriter (CorpusR)}
We build a simple word-level lookup rewriting model by exploiting the fully aligned words in {\APGC}~v2.1. We implement this model as a bigram maximum likelihood estimator: given an input word with its bigram surrounding context ($w_i$, $w_{i-1}$), a gender alternative target word ($y_i$), and a desired word-level target gender ($g$), the CorpusR model is built by computing $P(y_{i} | w_{i},w_{i-1}, g)$ over the training examples.
During inference, we generate all possible gender alternatives for the given input word ($w_i$). If the bigram context ($w_i$, $w_{i-1}$) was not observed in the training data, we backoff to a unigram context. If the input word was not observed during training, we pass it to the output as it is.

\paragraph{Morphological Rewriter (MorphR)}
For the morphological rewriter, we use the morphological analyzer and generator provided by CAMeL Tools \cite{obeid-etal-2020-camel}. We extend the
Standard Arabic Morphological Analyzer database (SAMA)~\cite{Graff:2009:standard} used by the morphological generator to produce controlled gender alternatives. 
%
We make our extensions to the database publicly available.\footnote{We provide code to reconstruct our extended database from the original SAMA 3.1 database (LDC2010L01) which can be obtained from the LDC.}
%
Given an input word and a desired word-level target gender, the morphological generator has the ability to produce gender alternatives by either rewriting the base word, its pronominal enclitics, or both. If an input word does not get recognized by the morphological analyzer and generator, we pass it to the output as it is. It is worth noting that this rewriting model does not require any training data.

\paragraph{Neural Rewriter (NeuralR)}
Inspired by work done on out-of-context morphological reinfection~\cite{kann-schutze-2016-single,cotterell-etal-2018-conll}, we design a character-level encoder-decoder model with attention.
Given an input word and word-level target gender, the encoder-decoder model would generate gender alternatives of the input word.
For the encoder, we use a two-layer bidrectional GRU~\cite{cho-etal-2014-learning} and for the decoder we use a two-layer GRU with additive attention~\cite{Bahdanau:2015:neural}. Furthermore, we employ \textit{side constraints}~\cite{sennrich-etal-2016-controlling} to control for the generation of gender alternatives. That is, we add the word-level target gender as a special token (e.g., <1F+B>) to the beginning of the input word and we feed that entire sequence to the model (i.e., <1F+B>\<سعيد>).
The intuition here is that the attentional encoder-decoder model would be able to learn to pay attention to the side constraints to generate the desired gender alternative of the input word. During inference, we use beam search to generate the top 3-best hypotheses.

\subsection{In-Context Ranking and Selection}
Since the three word-level gender alternative generation models we implement are out-of-context and given Arabic's morphological richness, we expect to get multiple output words when generating a single gender alternative for a particular input word. This leads to producing multiple candidate gender alternative output sentences. To select the best candidate output sentence, we rank all candidates in full sentential context based on their \textit{pseudo-log-likelihood} (PLL) scores~\cite{salazar-etal-2020-masked}. We first use Hugging Face's transformers to fine-tune the CAMeLBERT MSA model on the Input corpus of {\APGC}~v2.1 by using a masked language modeling \cite{devlin-etal-2019-bert} objective. This helps in mitigating the domain shift \cite{gretton-domain-shift} issue between CAMeLBERT's pretraining data and {\APGC}~v2.1. We then compute the PLL score for each sentence using the fine-tuned CAMeLBERT MSA model by masking the sentence tokens one by one.\footnote{We use \newcite{salazar-etal-2020-masked}'s implementation to compute the PLL scores: \url{https://github.com/awslabs/mlm-scoring}}
We will refer to the in-context ranking and selection as simply \textit{selection} throughout the paper.

Figure~\ref{fig:model_diag} presents an overview of our gender rewriting model. We describe the training settings and the model's hyperparameters in Appendix~\ref{sec:exp_setup}.

\section{Experimental Setup}
\label{sec:experiments}

\subsection{Evaluation Metrics}
We follow \newcite{alhafni-etal-2020-gender} by treating the gender rewriting problem as a user-aware grammatical error correction task and use the MaxMatch (M\textsuperscript{2}) scorer \cite{dahlmeier-ng-2012-better} as our evaluation metric.
The M\textsuperscript{2} scorer computes the precision (P), recall (R), and F\textsubscript{0.5} by maximally matching phrase-level edits made by a system to gold-standard edits. The gold edits are computed by the M\textsuperscript{2} scorer based on provided gold references.
Moreover and to be consistent with previous work, we also report BLEU \cite{Papineni:2002:bleu} scores which are obtained using SacreBLEU \cite{post-2018-call}. We report the gender rewriting results in a normalized space for Alif, Ya, and Ta-Marbuta \cite{Habash:2010:introduction}.
%

\subsection{Baselines}
\paragraph{Do Nothing} Our first baseline trivially passes the input sentences to the output as they are. This baseline highlights the level of similarity between the inputs and the outputs.

\paragraph{Joint Baseline Model} Our second baseline uses a  variant of the sentence-level linguistically enhanced joint gender identification and rewriting model introduced by \newcite{alhafni-etal-2020-gender}. 
The main difference between this model and the one introduced by \newcite{alhafni-etal-2020-gender} is that we model four multi-user target genders, whereas they only modeled two single-user target genders (1M, 1F). 
%
We implement this joint baseline model using a character-level GRU encoder-decoder with additive attention, where the learned input character-level representations are enriched with word-level morphological features obtained from the extended morphological analyzer that is part of CAMeL Tools.
The representation of the input sentence-level target gender (1M/2M, 1F/2M, 1M/2F, 1F/2F) is learned as part of the model and used during decoding when generating gender alternatives. We refer to this baseline as \textbf{Joint+Morph}. 

\paragraph{Extended Joint Baseline Models} Our third and fourth baseline models reduce the complexity of 
the \textbf{Joint+Morph} model
by not learning a representation for the input sentence-level target gender as part of the model. Instead, we provide the sentence-level target gender information as side constraints. We add the target gender as a special token to the beginning of the input sentence (e.g., <1M/2F>Input Sentence) when we feed it to the model. Moreover, we explore the effectiveness of enriching the input character representations with word-level morphological features. To do so, we omit the morphological features from the first joint variant, and we use them in the second. We refer to these models  as \textbf{Joint+Side Constraints} and \textbf{Joint+Morph+Side Constraints}, respectively.

\subsection{Our Multi-step Models} We explore five variants of the gender rewriting multi-step model described in \S\ref{sec:model_desc}. All five variants use the same gender identification (\textbf{GID}) and in-context selection models, but they differ in their out-of-context word-level gender rewriting generation setup. The first three variants use one word-level gender rewriting model each -- \textbf{CorpusR}, \textbf{MorphR}, or \textbf{NeuralR}.
%
%
The fourth multi-step model uses \textbf{MorphR} as a backoff if the input words that need to be rewritten are not observed by the \textbf{CorpusR} model during training ({\bf CorpusR>>MorphR}). 
The fifth system uses all three word-level gender alternative generation models in a backoff cascade:
{\bf CorpusR>>MorphR>>NeuralR}.

\input{results_table_dev_new}

\subsection{Data Augmentation}
\label{sec:aug-experiments}
Given the relatively small size of {\APGC}~v2.1 and motivated by recent work on using data augmentation to improve grammatical error correction \cite{wan-etal-2020-improving,stahlberg-kumar-2021-synthetic}, we investigate adding additional training examples through data augmentation. We randomly selected 800K sentences from the English-Arabic portion of the OpenSubtitles 2018 dataset, which was used to build {\APGC}. We ensured that all extracted pairs include either first or second (or both) person pronouns on the English side: \textit{I, me, my, mine, myself,} and \textit{you, your, yours, yourself}. 
To generate gender alternatives of the selected Arabic sentences, we pass each sentence four times through our best gender rewriting system to generate all four user gender contexts (1M/2M, 1F/2M, 1M/2F, 1F/2F).
We add the 800K selected Arabic sentences and their 1M/2M, 1F/2M, 1M/2F, 1F/2F generated gender alternatives to the Input, Target 1M/2M, Target 1F/2M, Target 1M/2F, and Target 1F/2F corpora of the training split of {\APGC}~v2.1, respectively. At the end, we end up with 857,603 Arabic parallel sentences (6,209,958 words). We make the synthetically generated data publicly available.


%

\input{results_table_test}

\section{Results}
\label{sec:results}
Table~\ref{tab:dev_results} presents the {\dev} set results. 
\textbf{Joint+Side Constraints} and \textbf{Joint+Morph+Side Constraints} 
significantly improve over the \textbf{Joint+Morph} baseline with up to 13.87 increase in F\textsubscript{0.5}.  
The best performing system overall is our multi-step model using all rewrite components -- Table~\ref{tab:dev_results}(i), henceforth, {\it Our Best Model}.
It improves over all the joint models in every compared metric, including a 22.84 increase in F\textsubscript{0.5} when compared to the \textbf{Joint+Morph} baseline.
{\it Our Best Model}'s biggest advantages seem to come from combining the three word-level out-of-context gender alternative generation models in a cascaded setup to deal with OOV words during the generation. Comparing (i) with (e,f,g) in Table~\ref{tab:dev_results}, we observe improvements ranging from 3.91 to 6.02 F\textsubscript{0.5}.

We used {\it Our Best Model} to conduct the data augmentation experiments as discussed in \S\ref{sec:aug-experiments}.  The full set of augmentation experiment results are presented in Appendix \ref{sec:augmentation_exps}. The best augmented model's results, which benefits from augmentation in the \textbf{GID} and \textbf{NeuralR} components, are also presented in Table~\ref{tab:dev_results}(j). However, its increase of 0.19 points in F\textsubscript{0.5} is not statistically significant with McNemar's~\cite{mcnemar1947} test at $p>0.05$.


The results of our best models on the {\test} sets of {\APGC}~v2.1 and {\APGC}~v1.0 are presented in Table~\ref{tab:test_results}. The results on {\APGC}~v2.1 {\test} show consistent conclusions with the {\dev} results. Our best augmented model improves over its non-augmented variant in every compared metric, including a 0.25 absolute increase in F\textsubscript{0.5} that is statistically significant with McNemar's test at $p<0.05$. For {\APGC}~v1.0, we do not report results with augmentation for fair comparison to previous results.  In both {\test} sets, we establish new SOTAs.




\input{error_analysis_table}
\input{results_table_MT_new}

\paragraph{Error Analysis}
We conducted an error analysis over the output of our best augmented system on {\APGC}~v2.1 {\dev}. In total, there were 1,475 (5.5\% out of 26,588) sentences with errors across the four target corpora. Table~\ref{tab:error_analysis} presents a summary of the error types our best augmented model makes.
%

The majority of errors (67.3\%) were caused by \textbf{GID} which achieves a word-level accuracy of 98.9\% on  {\dev}. The gender-rewriting errors constituted 18.1\% and selection errors 14.6\%.
Considering different target corpora, we observe that every time an F target is added, the number of errors increases. The 1M/2M target outputs has the lowest number of errors (268 or 18\%), while the 1M/2F targets outputs has the highest number of errors (480 or 33\%). 



\paragraph{Use Case: Post-Editing MT Output}
We demonstrate next how our proposed gender rewriting model could be used to personalize the output of user-unaware MT systems through post-editing. 
We use the English to Arabic Google Translate output sentences that are part of {\APGC}~v2.1. 
We evaluate Google Translate's output against all four target corpora (1M/2M, 1F/2M, 1M/2F, 1F/2F) separately. To re-target Google Translate's Arabic output for the four user gender contexts we model, we pass each Arabic sentence four times through our best augmented system (Table~\ref{tab:dev_results}(j)). We present the evaluation in terms of BLEU in Table~\ref{tab:MT_results} over {\APGC}~v2.1 {\test}. All the results are reported 
in an orthographically normalized space for Alif, Ya, and Ta-Marbuta.

Again, we observe that every time an M participant is switched to  F, the BLEU scores drop for Google Translate's output. This is consistent to what have been reported by \newcite{Alhafni:2022:corpus} and highlights the bias the machine translation output has towards masculine grammatical gender preferences. The post-edited outputs generated by our best augmented system improves over Google Translate's across the four target user contexts, achieving the highest increase in 2.27 BLEU points for 1F/2F. 

\section{Conclusion and Future Work}
\label{sec:conclusion}
We defined the task of gender rewriting in contexts involving two users (I and/or You), and developed a multi-step system that combines the positive aspects of both rule-based and neural rewriting models. 
Our best models establish the benchmark for this newly defined task and the SOTA for a previously defined first person version of it.
%
We further demonstrated a use case of our gender rewriting system by post-editing the output of a commercial MT system to provide personalized outputs based on the users' grammatical gender preferences.

In future work, we plan to explore the use of other pretrained models, and to work on the problem of gender rewriting in other languages and dialectal varieties.


\section*{Ethical Considerations}

\paragraph{Gender Rewriting} Our underlying intention of developing a gender rewriting model for Arabic is to increase the inclusiveness of NLP applications that deal with gender-marking morphologically rich languages. Our work aims at empowering and allowing users to interact with NLP technology in a way that is consistent with their social identities. We acknowledge that by limiting the choice of gender expressions to the grammatical gender choices in Arabic, we exclude other alternatives such as non-binary gender or no-gender expressions. However, we are not aware of any sociolinguistics published research that discusses such alternatives for Arabic. We stress on the importance of adapting Arabic NLP models to new gender alternative forms as they emerge as part of the language usage.
We further recognize the limitations of the gender identification component we use as part of our multi-step gender rewriting model as it relies on a language model that was pretrained on a large monolingual Arabic corpus, which could possibly contain biased text. We realize the potential risks of our proposed gender rewriting model if it is intentionally maliciously misused to produce gender alternatives that do not match the target users' gender preferences.


\paragraph{Data} We use the publicly available Arabic Parallel Gender Corpus ({\APGC}).\footnote{\url{http://resources.camel-lab.com/}} It is subject to its creators' own Copy Rights and User Agreement and we strictly adhere to its intended usage. It is worth noting that {\APGC} does not contain any heterocentric assumptions as part of its annotations (e.g., the word \<زوجي> `my husband' is labeled as gender-ambiguous (B)). Moreover, all proper names are labeled as B, even when they have strong gender-specific association \cite{Alhafni:2022:corpus}. The Arabic data we use for our data augmentation experiments was randomly sampled from OpenSubtitles 2018~\cite{Lison:2016:opensubtitles2016}, which is  publicly available.\footnote{\url{https://opus.nlpl.eu/OpenSubtitles-v2018.php}} OpenSubtitles is distributed
under a Creative Commons license.\footnote{Attribution-Non Commercial 4.0 International.}

\section*{Acknowledgements}
We thank Alberto Chierici, Christian Khairallah, Go Inoue, and Ossama Obeid for the helpful discussions. We acknowledge the support of the High Performance Computing Center at New York University Abu Dhabi. Finally, we wish to thank the anonymous reviewers
at NAACL 2022 for their feedback.

\bibliography{anthology,camel-bib-v2,extra}
\bibliographystyle{acl_natbib}

\clearpage

\appendix

\section{Detailed Experimental Setup}
\label{sec:exp_setup}

\paragraph{Gender Identification} We fine-tune CAMeLBERT MSA on a single GPU for 10 epochs with a learning rate of
5e-5, batch size of 32, a seed of 12345, and a maximum sequence length of 128. For the augmentation experiments, we use the same hyperparamters but we run the fine-tuning for 3 epochs. At the end of the fine-tuning, we pick the best checkpoint based on the performance on the {\dev} set. Our gender identification model has 108,506,901 parameters.

\paragraph{In-Context Ranking and Selection} We fine-tune CAMeLBERT MSA on a single GPU for 3 epochs with a learning of 5e-5, batch size of 32, and a seed of 88. The fine-tuned CAMeLBERT MSA model has 109,112,880 parameters.

\paragraph{Neural Rewriter (NeuralR)} For the character-level encoder-decoder neural rewriter model we use a character embedding size of 128, a hidden size of 256, a dropout probability of 0.2 on the outputs of each GRU layer, and gradient clipping with a maximum norm of 1. We train for 50 epochs on a single GPU with early stopping after 6 epochs if the loss does not decrease on the {\dev} set. We use the Adam \cite{kingma2014adam} optimizer with an initial learning rate of 5e-4, decaying by a factor of 0.5 if the loss on the {\dev} set does not decrease after 2 epochs. We train with greedy decoding and a batch size of 32. We also apply scheduled sampling (teacher forcing) \cite{Bengio:2015:scheduled} with a constant sampling probability (0.3) during training. During inference, we use beam search with a beam width of 10 to produce the top 3-best hypotheses. Our \textbf{NeuralR} model has 3,287,110 parameters.

\paragraph{Joint Models} The training settings and the hyperparameters of the joint models are identical to the ones we use in our \textbf{NeuralR} model.  

The crucial difference between the \textbf{Joint+Morph} model and its extended variants (\textbf{Joint+Side Constraints} and \textbf{Joint+Morph+Side Constraints}) is that the rewriting in the \textbf{Joint+Morph} model is conditioned on the sentence-level target gender. The representation of the sentence-level target gender in the baseline model is learned as an embedding of size 10 during training and only used in the decoder.

Our \textbf{Joint+Morph} model has 3,481,178 parameters; the \textbf{Joint+Side Constraints} model has 3,293,258 parameters; and the \textbf{Joint+Morph+Side Constraints} model has 3,480,926 parameters.

\paragraph{Training Time} The \textbf{CorpusR} model was trained on a single CPU and it took $\approx$2 minutes to be trained. All our neural models were trained on a single GPU. Fine-tuning CAMeLBERT MSA on the gender identification task took $\approx$1 hour; fine-tuning CAMeLBERT MSA on the MLM objective took $\approx$1 hour. Training the \textbf{NeuralR} model with different settings took $\approx$12 hours in total. All the baseline joint models took $\approx$29 hours to be trained.

It is worth noting that all the results presented in this work are reported over a single run and the hyperparameters of our neural models were manually tuned based on the performance on the {\dev} set.

\section{Arabic Parallel Gender Corpus v2.1: Extended Word-Level Annotations}
\label{sec:extended_word_labels}
\begin{table}[h!]
\centering
\includegraphics[width=.93\linewidth]{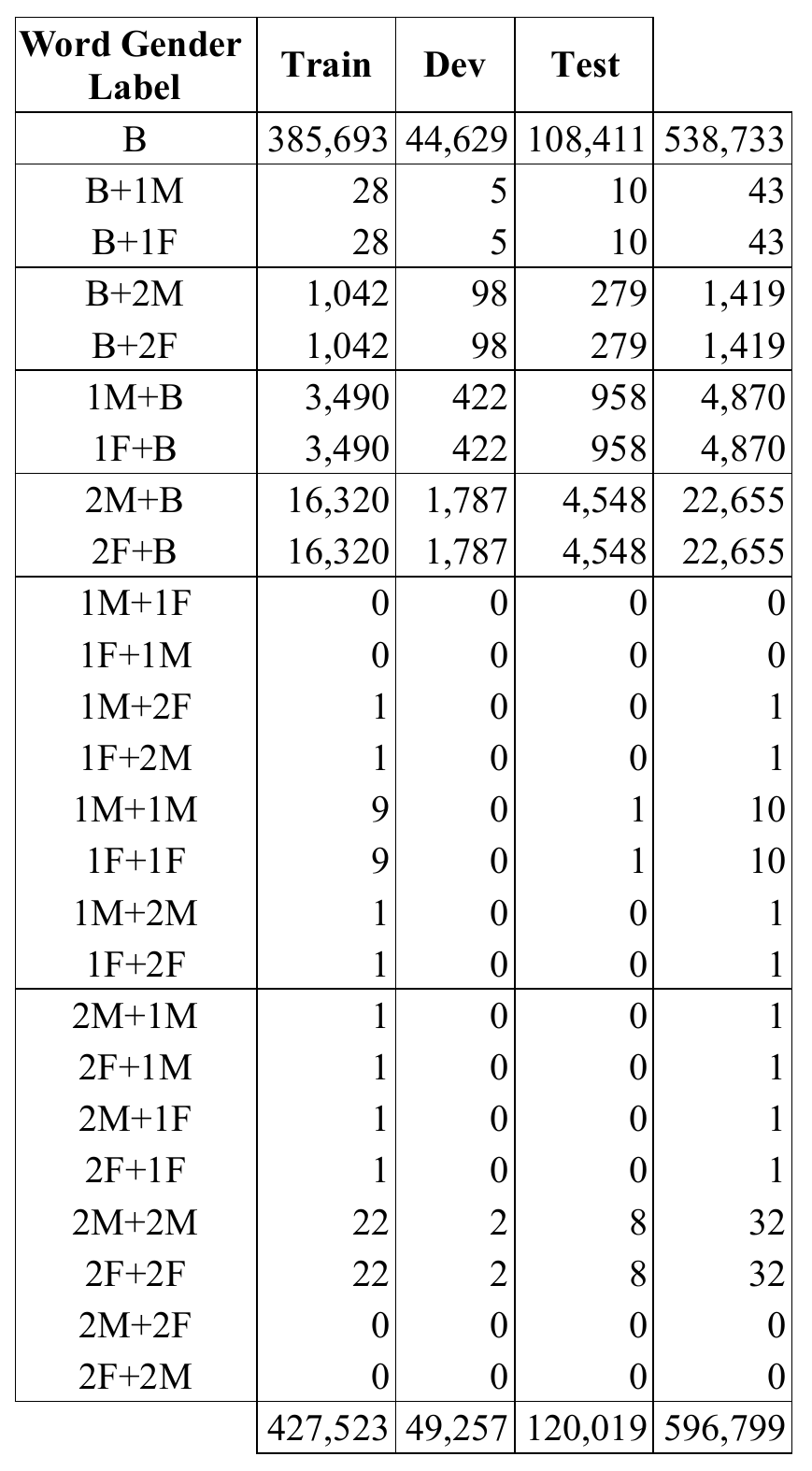}
\caption{The statistics of the extended word-level gender annotations of {\APGC}~v2.1 across the {\train}, {\dev}, and {\test} splits.}
\label{tab:word_stats_annotations}
\end{table}

\input{results_table_augmentation}

\section{Augmentation Experiments}
\label{sec:augmentation_exps}
When it comes to the data augmentation experiments, we took the best performing system (Table \ref{tab:dev_results}(i)) and explored training its different components on the augmented training data. Evaluation results on the {\dev} set of {\APGC}~v2.1 using data augmentation are presented in Table~\ref{tab:dev_aug_results}. 
Starting off with training the \textbf{CorpusR} model on the augmented data (Table~\ref{tab:dev_aug_results}(a)), we notice a decrease in performance by 0.23 F\textsubscript{0.5} compared to {\it Our Best Model} (Table~\ref{tab:dev_results}(i)). This is attributed to the noisy coverage increase in the \textbf{CorpusR} model and can be observed by the decrease in precision (88.19) and recall (86.66). 
When we train the \textbf{GID} and the \textbf{NeuralR} models on the augmented data (Table~\ref{tab:dev_aug_results}(b-c)), we get an increase in F\textsubscript{0.5} reaching 88.28 and 88.12, respectively. However, using both the augmented \textbf{GID} and \textbf{CorpusR} models (Table~\ref{tab:dev_aug_results}(d)) decreases the performance slightly as it achieves 88.10 in F\textsubscript{0.5}. The best performing system was the one that uses both the augmented \textbf{GID}  and \textbf{NeuralR} models (Table~\ref{tab:dev_aug_results}(e)) as it improves over its non-augmented variant reaching 88.30 (0.19 increase) in  F\textsubscript{0.5}. Combining the three augmented \textbf{GID}, \textbf{CorpusR}, and \textbf{NeuralR} models (Table~\ref{tab:dev_aug_results}(f)) achieves 88.08 in F\textsubscript{0.5}.

\end{document}

%% file: macros.tex
\newcommand{\hide}[1]{}

\newcommand{\AYN}{{$\varsigma$}}

%% file: results_table_dev_new.tex
\tabcolsep=2.5pt
\begin{table*}[ht!]
        \centering
        \begin{tabular}{|llllllllll|c|c|c|c|}
            \cline{11-14}

            \multicolumn{10}{l|}{} & \textbf{P}  & \textbf{R} & \textbf{F\textsubscript{0.5}} & \textbf{BLEU} \\\cline{1-14}
            \multicolumn{1}{|l}{(a)} & \multicolumn{9}{l|}{\textbf{Do Nothing}} & 100.0 & 0.0 & 0.0 & 89.36 \\\cline{1-14}
             \multicolumn{1}{|l}{(b)} &\multicolumn{1}{l}{\textbf{Joint}} & + & \multicolumn{7}{l|}{\textbf{Morph}} & 64.76 &	67.40 & 65.27 & 93.31 \\\hline\hline
            
            \multicolumn{1}{|l}{(c)} &\multicolumn{1}{l}{\textbf{Joint}} & + & \multicolumn{7}{l|}{\textbf{Side Constraints}} &  77.10 & 77.71 & 77.22 & 95.60 \\\cline{1-14}
            
            \multicolumn{1}{|l}{(d)} &\multicolumn{1}{l}{\textbf{Joint}} & + & \multicolumn{1}{l}{\textbf{Morph}} & + & \multicolumn{5}{l|}{\textbf{Side Constraints}} &  78.97 & 79.84 & 79.14 & 96.17 \\\hline\hline
            
            \multicolumn{1}{|l}{(e)} &\multicolumn{1}{l}{\textbf{GID}} & + & \multicolumn{1}{l}{\textbf{CorpusR}} & + & \multicolumn{5}{l|}{\textbf{Selection}}  &  88.22 & 71.22 & 84.20 & 96.54 \\\cline{1-14}
            
            \multicolumn{1}{|l}{(f)} &\multicolumn{1}{l}{\textbf{GID}} & + & \multicolumn{1}{l}{\textbf{MorphR}} & + & \multicolumn{5}{l|}{\textbf{Selection}} & 84.48 & 75.29 & 	82.47 & 96.96  \\\cline{1-14}
            
            \multicolumn{1}{|l}{(g)} &\multicolumn{1}{l}{\textbf{GID}} & + & \multicolumn{1}{l}{\textbf{NeuralR}} & + & \multicolumn{5}{l|}{\textbf{Selection}}& 84.62 & 73.32 & 82.09 & 96.75 \\\hline\hline
            
            \multicolumn{1}{|l}{(h)} &\multicolumn{1}{l}{\textbf{GID}} & + & \multicolumn{1}{l}{\textbf{CorpusR}} & >> &
            \multicolumn{1}{l}{\textbf{MorphR}} & + & \multicolumn{3}{l|}{\textbf{Selection}} &
            \textbf{88.59} & 85.84 & 88.02 & 97.96 \\\cline{1-14}
            
            \multicolumn{1}{|l}{(i)} & \multicolumn{1}{l}{\textbf{GID}} & + & \multicolumn{1}{l}{\textbf{CorpusR}} & >> &
            \multicolumn{1}{l}{\textbf{MorphR}} & >> & \multicolumn{1}{l}{\textbf{NeuralR}} & + & \multicolumn{1}{l|}{\textbf{Selection}} & 88.46 & \bf 86.74 & \bf 88.11 & \bf 98.04 \\\hline\hline
\multicolumn{1}{|l}{(j)} & \multicolumn{1}{l}{\textbf{GID\textsubscript{Aug}}} & + & \multicolumn{1}{l}{\textbf{CorpusR}} & >> &
\multicolumn{1}{l}{\textbf{MorphR}} & >> & \multicolumn{1}{l}{\textbf{NeuralR\textsubscript{Aug}}} & + & \multicolumn{1}{l|}{\textbf{Selection}}& \bf 88.67 & \bf 86.84 & \bf 88.30 & \bf 98.05 \\\cline{1-14}

\end{tabular}
\caption{Results of a number of systems on the {\dev} set of {\APGC} v2.1.
\textbf{Aug}  indicates using augmented data.}
\label{tab:dev_results}
\end{table*}

%% file: results_table_test.tex
\tabcolsep=2.5pt
\begin{table*}[ht!]
        \centering
        \begin{tabular}{|c|lllllllll|c|c|c|c|}
            \cline{11-14}
    \multicolumn{1}{l}{}  & \multicolumn{9}{l|}{} & \textbf{P}  & \textbf{R} & \textbf{F\textsubscript{0.5}} & \textbf{BLEU} \\\cline{1-14}
            
       \bf APGC  & \multicolumn{1}{|l}{\textbf{Joint}} & + & \multicolumn{1}{l}{\textbf{Morph}} & + & \multicolumn{5}{l|}{\textbf{Side Constraints}} & 79.27 & 80.44 & 79.50  & 96.19 \\\cline{2-14}
            
       \bf v2.1    & \multicolumn{1}{|l}{\textbf{GID}} & + & \multicolumn{1}{l}{\textbf{CorpusR}} & >> &
            \multicolumn{1}{l}{\textbf{MorphR}} & >> & \multicolumn{1}{l}{\textbf{NeuralR}} & + & \multicolumn{1}{l|}{\textbf{Selection}} & 88.70	& 86.13 & 88.17 & 97.98 \\\cline{2-14}
            
        \bf  Test   & \multicolumn{1}{|l}{\textbf{GID\textsubscript{Aug}}} & + & \multicolumn{1}{l}{\textbf{CorpusR}} & >> &
            \multicolumn{1}{l}{\textbf{MorphR}} & >> & \multicolumn{1}{l}{\textbf{NeuralR\textsubscript{Aug}}} & + & \multicolumn{1}{l|}{\textbf{Selection}} & \bf 88.86 & \bf 86.69 & \bf 88.42 & \bf 98.05 \\\hline\hline

        \bf APGC    &  \multicolumn{9}{|l|}{\textbf{\newcite{habash-etal-2019-automatic}}}&  77.74 & 52.06 & 70.76 & 98.28 \\\cline{2-14}
      \bf v1.0    &   \multicolumn{9}{|l|}{\textbf{\newcite{alhafni-etal-2020-gender}}}&  \textbf{78.98} & 60.32 & 74.38 & 98.49 \\\cline{2-14}
       \bf  Test     &  \multicolumn{1}{|l}{\textbf{GID}} & + & \multicolumn{1}{l}{\textbf{CorpusR}} & >> &
            \multicolumn{1}{l}{\textbf{MorphR}} & >> & \multicolumn{1}{l}{\textbf{NeuralR}} & + & \multicolumn{1}{l|}{\textbf{Selection}} & 78.57 & \textbf{73.17} & \textbf{77.43} & \textbf{98.92} \\\hline

        \end{tabular}
        \caption{Gender rewriting results on the {\test} sets of {\APGC} v2.1 and {\APGC} v1.0.}
        \label{tab:test_results}
\end{table*}

%% file: error_analysis_table.tex
\begin{table}[t]
\centering
\setlength{\tabcolsep}{1.5pt}
\begin{tabular}{l|rr|rr|rr|rr|}
\cline{2-9}
          & \multicolumn{2}{c|}{\bf 1M/2M} & \multicolumn{2}{c|}{\bf 1F/2M} & \multicolumn{2}{c|}{\bf 1M/2F} & \multicolumn{2}{c|}{\bf 1F/2F} \\\hline\hline
\multicolumn{1}{|l|}{\bf GID} & 150        & 56\%         & 194        & 70\%         & 325        & 68\%         & 324        & 72\%         \\\hline
\multicolumn{1}{|l|}{\bf Rewrite} & 69         & 26\%         & 50        & 18\%         & 82        & 17\%         & 66        & 15\%         \\\hline
\multicolumn{1}{|l|}{\bf Select} & 49         & 18\%         & 35        & 13\%         & 73         & 15\%         & 58         & 13\%         \\\hline
   \multicolumn{1}{|c|}{\it Total}      & \multicolumn{2}{c|}{268}        & \multicolumn{2}{c|}{279}        & \multicolumn{2}{c|}{480 }       & \multicolumn{2}{c|}{448}           \\\hline  
\end{tabular}
\caption{Error type statistics of our best augmented system's performance on {\APGC} v2.1 {\dev}.}
\label{tab:error_analysis}
\end{table}

%% file: results_table_MT_new.tex
\begin{table}[ht!]
\tabcolsep=1pt
        \centering
        \begin{tabular}{|l|c|c|c|c|}
            \cline{1-5}
            {\textbf{Target}} & \textbf{1M/2M}  & \textbf{1F/2M} & \textbf{1M/2F} & \textbf{1F/2F} \\\hline\hline
            {\textbf {Google Translate}} & 13.59 & 13.15 & 11.38 & 10.96 \\\hline\hline
            
            \textbf{Best System\textsubscript{Aug}} & \textbf{13.71} & \textbf{13.64} & \textbf{13.30}	& \textbf{13.23}\\\hline

        \end{tabular}
        \caption{BLEU results on the post-edited Google Translate output of {\APGC} v2.1 {\test} using our best augmented system.}
        \label{tab:MT_results}
    \end{table}

\hide{\tabcolsep=3pt
\begin{table*}[ht!]
        \centering
        \begin{tabular}{|l|c|c|c|c||c|c|c|c|}
            \cline{2-9}
            \multicolumn{1}{l|}{}  &  \multicolumn{4}{c||}{{\dev}}  &  \multicolumn{4}{c|}{{\test}} \\\cline{1-9}
            {\textbf{Target}} & \textbf{1M/2M}  & \textbf{1F/2M} & \textbf{1M/2F} & \textbf{1F/2F} & \textbf{1M/2M}  & \textbf{1F/2M} & \textbf{1M/2F} & \textbf{1F/2F} \\\hline\hline
            {\textbf {Google Translate}} &  13.8 & 13.3 & 11.7 & 11.3 & 
            13.6 & 13.2 & 11.4 & 11.0 \\\hline\hline
            
            \textbf{Best System (Aug)} &  \textbf{13.8} & \textbf{13.8} & \textbf{13.5} & \textbf{13.5} & \textbf{13.7} & \textbf{13.7} & \textbf{13.4}	& \textbf{13.3}\\\hline

        \end{tabular}
        \caption{BLEU results on the post-edited Google Translate output of the {\dev} set of {\APGC} v2.0.}
        \label{tab:MT_results}
    \end{table*}
}

%% file: results_table_augmentation.tex
\tabcolsep=3pt
\begin{table*}[ht!]
        \centering
        \begin{tabular}{|llllllllll|c|c|c|c|}
            \cline{11-14}

            \multicolumn{10}{l|}{} & \textbf{P}  & \textbf{R} & \textbf{F\textsubscript{0.5}} & \textbf{BLEU} \\\cline{1-14}

            \multicolumn{1}{|l}{(a)} & \multicolumn{1}{l}{\textbf{GID}} & + & \multicolumn{1}{l}{\textbf{CorpusR\textsubscript{Aug}}} & >> &
            \multicolumn{1}{l}{\textbf{MorphR}} & >> & \multicolumn{1}{l}{\textbf{NeuralR}} & + & \multicolumn{1}{l|}{\textbf{Selection}} & 88.19 & 86.66 & 87.88 & 98.04 \\\cline{1-14}
            
            \multicolumn{1}{|l}{(b)} &\multicolumn{1}{l}{\textbf{GID\textsubscript{Aug}}} & + & \multicolumn{1}{l}{\textbf{CorpusR}} & >> &
            \multicolumn{1}{l}{\textbf{MorphR}} & >> & \multicolumn{1}{l}{\textbf{NeuralR}} & + & \multicolumn{1}{l|}{\textbf{Selection}} & 88.63 & 	\bf 86.91 & 88.28 & 98.05 \\\cline{1-14}
            
            \multicolumn{1}{|l}{(c)} & \multicolumn{1}{l}{\textbf{GID}} & + & \multicolumn{1}{l}{\textbf{CorpusR}} & >> &
            \multicolumn{1}{l}{\textbf{MorphR}} & >> & \multicolumn{1}{l}{\textbf{NeuralR\textsubscript{Aug}}} & + & \multicolumn{1}{l|}{\textbf{Selection}}& 88.50 & 86.64 & 88.12 & 98.04 \\\cline{1-14}
            
            \multicolumn{1}{|l}{(d)} & \multicolumn{1}{l}{\textbf{GID\textsubscript{Aug}}} & + & \multicolumn{1}{l}{\textbf{CorpusR\textsubscript{Aug}}} & >> &
            \multicolumn{1}{l}{\textbf{MorphR}} & >> & \multicolumn{1}{l}{\textbf{NeuralR}} & + & \multicolumn{1}{l|}{\textbf{Selection}}& 88.41 & 86.89 & 88.10 & \bf 98.06 \\\cline{1-14}
            
            \multicolumn{1}{|l}{(e)} & \multicolumn{1}{l}{\textbf{GID\textsubscript{Aug}}} & + & \multicolumn{1}{l}{\textbf{CorpusR}} & >> &
            \multicolumn{1}{l}{\textbf{MorphR}} & >> & \multicolumn{1}{l}{\textbf{NeuralR\textsubscript{Aug}}} & + & \multicolumn{1}{l|}{\textbf{Selection}}& \bf 88.67 & 86.84 & \bf 88.30 & 98.05 \\\cline{1-14}
            
            \multicolumn{1}{|l}{(f)} & \multicolumn{1}{l}{\textbf{GID\textsubscript{Aug}}} & + & \multicolumn{1}{l}{\textbf{CorpusR\textsubscript{Aug}}} & >> &
            \multicolumn{1}{l}{\textbf{MorphR}} & >> & \multicolumn{1}{l}{\textbf{NeuralR\textsubscript{Aug}}} & + & \multicolumn{1}{l|}{\textbf{Selection}} & 88.39 &	86.87 & 88.08 & 98.05 \\\cline{1-14}

\end{tabular}
\caption{Results of the data augmentation experiments on the {\dev} set of {\APGC} v2.1.
\textbf{Aug} indicates that the component of the system is trained on the augmented data.}
\label{tab:dev_aug_results}
\end{table*}